\documentclass{article}

    \usepackage[preprint, nonatbib]{neurips_2020}

\usepackage[utf8]{inputenc} 
\usepackage[T1]{fontenc}    
\usepackage{hyperref}       
\usepackage{url}            
\usepackage{booktabs}       
\usepackage{amsfonts}       
\usepackage{nicefrac}       
\usepackage{microtype}      
\usepackage{amsmath}        
\usepackage{amssymb}        
\usepackage{graphicx}       
\usepackage{subcaption}     
\usepackage[export]{adjustbox} 
\usepackage{color}

\title{BHN: A Brain-like Heterogeneous Network}

\author{%
    Liu, Tao \\
}

\begin{document}

\maketitle

\begin{abstract}
The human brain works in an unsupervised way, and more than one brain region is essential for lighting up intelligence.
Inspired by this, we propose a brain-like heterogeneous network (BHN), which can cooperatively learn a lot of distributed representations and one global attention representation.
By optimizing distributed, self-supervised, and gradient-isolated objective functions in a minimax fashion, our model improves its representations, which are generated from patches of pictures or frames of videos in experiments.
\end{abstract}

\section{Introduction}
\label{Introduction}
It is a mystery how different brain regions to be optimized jointly.
In this article, we propose a brain-like heterogeneous network(BHN) simulating the multi-module structure of the brain.

We use three hypothesises in this article:
\begin{enumerate}
 \item The brain is a machine to maximize information of its inner representations. This hypothesis was known as Efficient Coding\cite{barlow1961possible} or Efficient Information Representation\cite{linsker1990perceptual,atick1992could}.
 \item The brain learns by optimizing certain objective functions, and different brain regions optimize different objective functions\cite{lake2017building}.
 \item The brain works by fusing top-down predictions with bottom-up perceptions. This hypothesis actually enables the brain to process information recursively.
\end{enumerate}

We view hypothesis 1 as the first principle to understand the brain and obtain desired objective functions by formalizing it. The objective is a sum of many objective functions, each applied on an individual module, and all the modules make up BHN. 
We also seek to understand the brain's information processing scheme, which we name as Recursive Modeling in this article.

Following in this article, firstly, the section \ref{Info} will give the objective functions derived from the first hypothesis.
Next, the section \ref{BHN} and the section \ref{RM} will elaborate on BHN and Recursive Modeling respectively.
And then the section \ref{A} and the section \ref{B} will provide some demonstration experiments for the  former two sections respectively.

\section{Efficient Information Representation}
\label{Info}
The brain collects information from the environment($x$) and then generates internal representations($z$).
It is inferred that an important function of the brain is to maximize the information entropy of its representations.
It is generally believed that these representations are distributed on the cerebral cortex, and so it is essential to ensure the independence of information they represent.
Previous solutions include sparse-coding\cite{olshausen1996emergence}, independent component analysis\cite{hyvarinen2000independent}, and end-to-end deep learning.
In this article we propose our solution as follows.

We use$\{z^{1}, z^{2}, \cdots, z^{n}\}$ to denote the representations distributed on the cerebral cortex, and use $H(z^{1} z^{2} \cdots z^{n})$ to denote the information entropy of them.
We then formalize the objective function as $\max H(z^{1} z^{2} \cdots z^{n})$. 

Considering 
\begin{equation}
    H(z^{1} z^{2} \cdots z^{n}) =\sum_iH(z^{i}) +[H(z^{1} z^{2} \cdots z^{n})-\sum_iH(z^{i})]
\end{equation}
the objective function can be roughly decomposed into two sub-objectives\cite{atick1992could}, as
\begin{equation}
\begin{cases}
 \max\limits_{z}\ H(z^{i})  \\
 \min\limits_{z}\ I(z^{i};z^{j}),   & \mbox{if } i \ne j
\end{cases}
    \label{eq:minmax}
\end{equation}
Noting that the second sub-objective is intractable because of the $\Omega (n^2)$ computational complexity, so we introduce a \textbf{global} attention\cite{graves2014neural,vaswani2017attention} representation($a$) into (\ref{eq:minmax}) by reforming the expression in a minimax fashion, as
\begin{equation}
\begin{cases}
&\max\limits_{z}\ \sum_iH(z^{i})
\\
&\min\limits_{z}\max\limits_{a}\ \sum_i I(z^{i};a)
\end{cases}
\label{eq:minmax_2}
\end{equation}
Then, by re-composing the two expressions above into a single one, we obtain the objective function: 
\begin{equation}
\label{eq:minz}
    \min\limits_{z}\max\limits_{a} \sum_i [-H(z^{i}) + I(a;z^{i})]
\end{equation}

We use contrastive losses\cite{hadsell2006dimensionality} to formalize $H(z^i)$. Contrastive losses measure the similarities of sample pairs in a representation space. A form of a contrastive loss function, which is called InfoNCE\cite{oord2018representation}, is considered in this article:
\begin{equation}
\label{eq:hzi}
    H(z^{i}) \propto \log
\frac{\exp(f(z^{i}, z^{i}_{+}))}{\sum_{z^{i}_{-} \in Z^{i}} \exp(f(z^{i}, z^{i}_{-}))}
\end{equation}
where $f$ is a density ratio, which preserves the mutual information between a positive or negative pair of samples.

The next step is to formalize $I(a;z^{i})$.
To stabilize minimaxing on $I(a;z^{i})$, we do not formalize it directly. Instead, we use $a$ to produce a probability distribution, i.e. $P(z^i)$, as the prediction of $z^{i}$. The $a$ is called as an attention representation because it is used to generate shared Query/Key vectors $a^i$, each of which is paired with a representation $z^i$, and these Q/K vectors will be used to calculated each sample's probability/weight.
The details are as follows:

We provide a memory pool having $N$ paired samples, as
\begin{equation}
X^{i}=(A^{i},Z^{i})=\{(a^{i}_1,z^{i}_1),(a^{i}_2,z^{i}_2),\cdots,\cdots, (a^{i}_N,z^{i}_N)\}
\end{equation}
where $Z^{i}$ is the sample space of $P(z^i)$. The probability $P\big|z^i = z_j^i$ is equal to the attention weight $w_j^i$ calculated by
\begin{equation}
P\big|_{z^i = z_j^i}
 =w_j^i
 =
    \text{softmax}\left(
    {similarity}( a^i ,a_j^i)
    \right)
    \bigg|_{a^{i}_{j} \in A^{i}}
\end{equation}

Now we can formalize $I(a;z^{i})$ as
\begin{equation}
\label{eq:iaz}
I(a;z^{i}) \propto \log
\frac{\sum_{z^{i}_{j} \in Z^{i}} w_j^i \exp(f(z^{i}, z^{i}_{j}))}{\sum_{z^{i}_{-} \in Z^{i}} \exp(f(z^{i}, z^{i}_{-}))} 
\end{equation}

Eventually, by bringing (\ref{eq:hzi}) and (\ref{eq:iaz}) into (\ref{eq:minz}), we formalize the objective function as
\begin{equation}
\label{eq:obj}
    \min\limits_{z}\max\limits_{a}
    \sum_i[-\log \frac{\exp(f(z^{i}, z^{i}_{+}))}
    {\sum_{z^{i}_{j} \in Z^{i}} w_j^i \exp(f(z^{i}, z^{i}_{j}))}]
\end{equation}

Notable, this objective function suggests a probabilistic inference machines\cite{von2013treatise} and it is the corollary of our hypothesises. This is biologically plausible and we can say that \textbf{the attention representation makes predictions by activating selective replays of representations in the cerebral cortex}.

\section{Brain-like Heterogeneous Network}
\label{BHN}
In the section we propose the architecture of BHN to apply the objective function in Equation \ref{eq:obj}. It has a cortex-network composed of basic units, with unit $i$ generating the corticocerebral representation $z^i$, and an attention-network generating the global attention representation $a$.
As the name suggests, we use artificial neural network(ANN) to implement these two components.
Different from popular approach using end-to-end back-propagation with a global loss function, in our model, there is gradient isolation between the units and between the two networks.

\begin{figure}[ht]
  \centering
  \includegraphics[width=0.9\textwidth]{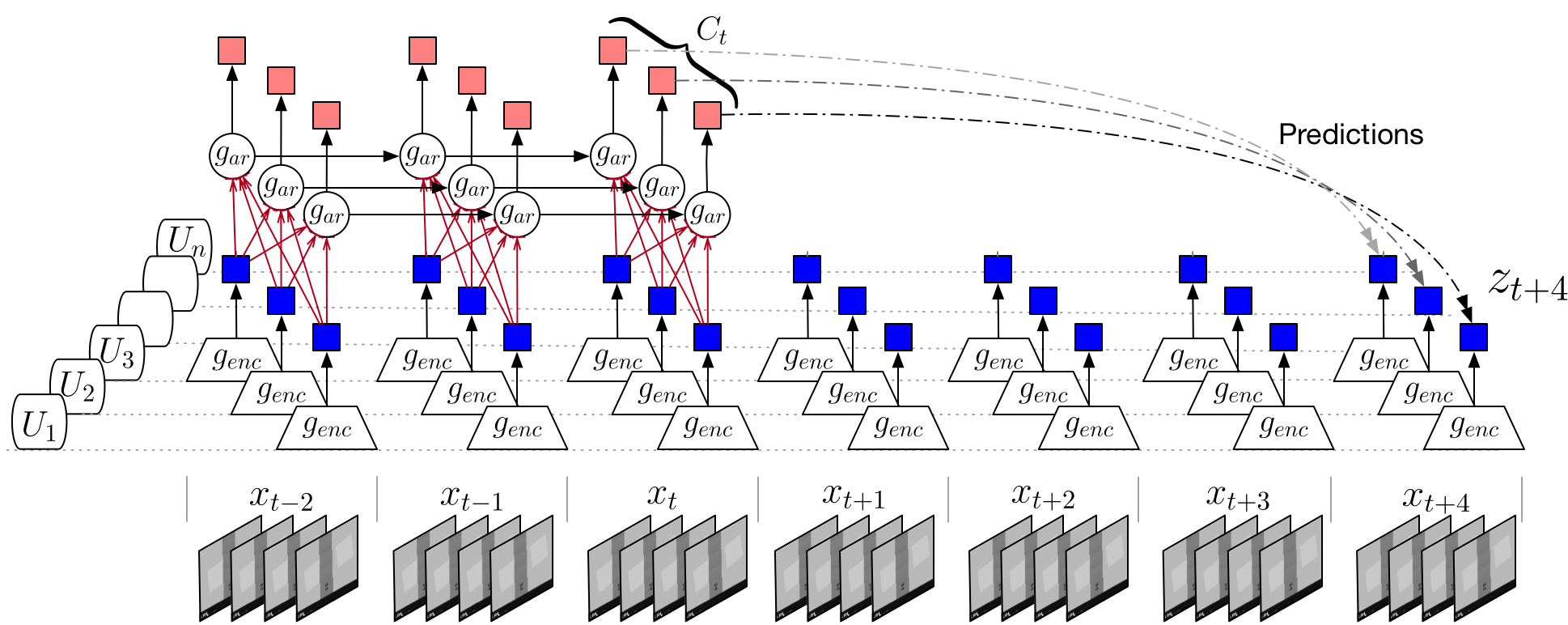}
  \caption{\label{fig:cortex} Architecture of the cortex-network in video tasks}
\end{figure}

\paragraph{Cortex-network} In each unit $i$, there is an encoder $g_{enc}^i$ to encode the input $x$ into a latent representation $z^i$. In the following image task, there only is a $g_{enc}^i$ inside the unit. While in the video tasks, in each unit, another network, which is called aggregator $g_{ar}^i$, is used to output a unit context $c^i$ to act as the positive partner of the $z^i$. Actually, we are applying Contrastive Predictive Coding\cite{oord2018representation} in each unit, as shown in Figure \ref{fig:cortex}.

\begin{figure}[hb]
  \centering
  \begin{subfigure}{0.35\textwidth}
    \includegraphics[width=0.7\linewidth]{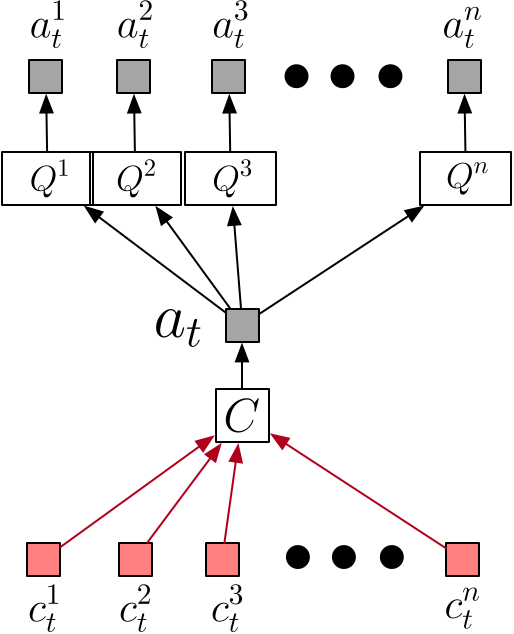} 
    \caption{}
    \label{fig:h1}
    \end{subfigure}
    \begin{subfigure}{0.35\textwidth}
    \includegraphics[width=0.7\linewidth, right]{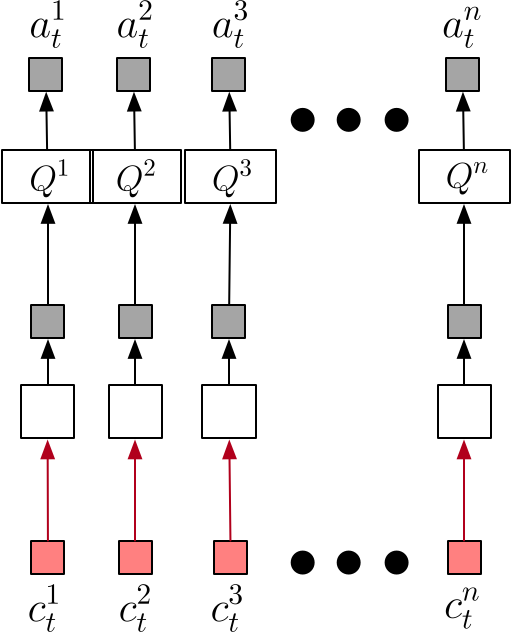}
    \caption{control group 2}
    \label{fig:h2}
    \end{subfigure}
  \caption{\label{fig:hippocampus} Architecture of the attention-network in video tasks}
\end{figure}

\paragraph{Attention-network}
The attention-network generates the global attention representation $a$, like the medial temporal lobe in the mammalian brain.
Its architecture is like a traditional encoder-decoder network, where the encoder generates $a$ and the the decoder generates $a^i$, as shown in Figure \ref{fig:h1}.

The input of the attention-network is from the output of the cortex-network. In our image task, it is natural to take all $z^i$ as input because they are the only outputs of the cortex-network. In our video tasks, the attention-network takes all $c^i$ as input, because we want to get $a$ in advance of $z^i$.

The attention representation $a$ should not retain all the information inputted into it, but only needs to capture the information shared by multiple units.
To achieve this goal, one option is to make $a$ act as an information bottleneck, which means that $a$ is lower dimensional than the input vector.
The other option is to arbitrarily drop out some units' outputs. In our image task, we adopt the second option, and in our video tasks, we adopt the first one.

\paragraph{Neural Interface} Neural interface is not an essential component of BHN, but we want to mention it here in advance because it is important for the Recursive Modeling. Unlike the cortex-network and the attention-network, Neural Interfaces have no biological counterparts.
Actually, this name comes from Brain-Computer Interfaces(BCIs) \cite{wolpaw2000brain}.
By processing information from the cortex-network, neural interfaces perform various functions, such as controlling attention, controlling actions, and whatever as you need.

\section{Experiment(1)}
\label{A}
\subsection{Image Task}
We download ten landscape pictures from the internet and crop them into 8000 patches of $\ 16 \times 16 $ pixels, and then each patch is gray-scaled and normalized.
We then design a BHN model to learn on this dataset.
The model has 64 units in its cortex-network.
The encoder in each unit contains 128 hidden units with leaky-relu activation, and the attention-network contains 256 hidden units, which is also the dimension of $a$, with leaky-relu activation.
The dimensions of $z^i$ and $a^i$ are both set to 1.
The batch size, which is also the size of $X^{i}$, is 512.

The density ratio $f$ is formalized as
\begin{equation}
    f(z, \mathfrak{z}) = - \text{clamp\_max\_5}(|z- \mathfrak{z}|)
\end{equation}

The similarity between $a^i$ is formalized as
\begin{equation}
    {similarity}(a^i ,a_j^i) = -|a^i-a_j^i|/\tau
\end{equation}
where $\tau$ is the temperature optimized together with the attention-network.

The inputs are added with Gaussian white noise with $mean=0$ and $std=0.1$ for image enhancement, and also in this way to produce positive sample pairs in constrastive loss function.
The dropout ratio is $0.2$ in the attention-network.
We use SGD optimizer with the $lr=0.1$, $momentum=0.9$, and $weight\_decay=0.001$. The model is light and the training runs fast even in a laptop without GPU acceleration. 

\begin{figure}[hb]
  \centering
    \begin{subfigure}{0.32\textwidth}
    \includegraphics[width=1\linewidth]{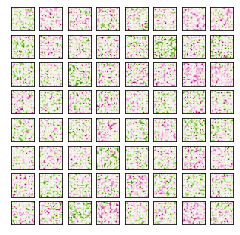} 
    \caption{untrained}
    \label{fig:v0}
    \end{subfigure}
    \begin{subfigure}{0.32\textwidth}
    \includegraphics[width=1\linewidth]{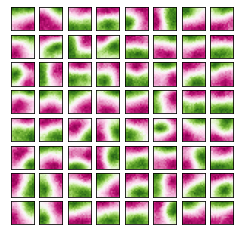}
    \caption{normal}
    \label{fig:v2}
    \end{subfigure}
    \begin{subfigure}{0.32\textwidth}
    \includegraphics[width=1\linewidth]{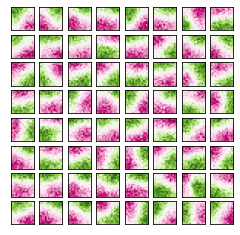} 
    \caption{control}
    \label{fig:v1}
    \end{subfigure}

  \caption{\label{fig:vision} Visualized features of units}
\end{figure}

In addition to the normal experiment, we also establish a control experiment where the objective function is to $\max \sum_i H(z^i)$ only. 
After 40 epochs of training, we visualize all 64 units by maximizing their outputs.
The results are shown in Figure \ref{fig:vision}.

In order to show clearly, we use red and green to indicate light and shade.
As can be seen from the figure, the visualized features are noisy if the model is not trained. 
In both normal and control experiments, the units have intensified responses to certain image modes after training.
The result images in the control experiment are fuzzy.
In contrast, the visualized features in the normal experiment are more sharp and diverse.

\subsection{Video Task}
\label{A:v}
We build a video set containing 64 episodes recording the play of CarRacing game in OpenAI gym. Each episode lasts for 512 frames and each frame has a size of (96, 96) pixels. The frames are converted to gray scale and rescaled to (-1,1).
At each time step, 4 consecutive frames with additional noises are fed to the input.  

A linear layer, shared by all $g^{i}_{enc}$ for the consideration of reducing the number of parameters, will first reduce the dimensions of inputs from $ 4 \times (96\times 96)$ to $512$.
The encoder architecture $g^{i}_{enc}$ contains 32 hidden units with leaky-relu activations.
We then use a GRU-RNN \cite{cho2014learning} for the autoregressive part of the unit, $g^{i}_{ar}$, with 32 dimensional hidden state. 
The cortex-network has 16 units, and the attention-network is a simple unbiased linear network with a hidden layer. Dimensions of $z^{i}_t$, $c^{i}_t$ , $a^{i}_t$ and $a_t$ are all set to 2.
The batch size, which is also the size of $X^{i}$, is 256.

In our experiment, $z_{t+4}^i$ and $c_{t}^i$ are used as the positive pair for the contrastive loss function. The delay of $4$ is, somewhat arbitrary, to quantify the directional information between $z$ and $c$.

The density ratio $f$ is formalized as
\begin{equation}
    f(z^i_{t+4}, c^i_t) = -\cos\langle z^i_{t+4},c^i_t\rangle/T
\end{equation}
where $T$ is the temperature optimized together with the cortex-network.

The similarity between $a^i$ is formalized as
\begin{equation}
    {similarity}(a^i ,a_j^i) = -\cos\langle a^i, a_j^i\rangle/\tau
\end{equation}
where $\tau$ is the temperature optimized together with the attention-network.

We choose Adam optimizer with $lr=1e-4$. We use data enhancement in which each episode is folded into 16 segments of 256 frames long. We train each model for 20 epochs. However, according to our experience, a much longer training would not lead to over-fitting.

We use deconvolutional networks\cite{zeiler2011adaptive} to reconstruct images from representations $z_t$, $c_t$ and $a_t$ respectively.
Mean square errors($mse$) of the reconstructed images will be used to evaluate the quality of source representations.
Given that a trivial solution could achieve a loss of 0.0225 if none information was provided, in the following, we use the score, calculated by $(0.0225-{mse}) \times 255$, to indicate the quality.

We also establish two control groups to demonstrate the performance of adversarial training.
\paragraph{control groups 1}We abandon the attention-network to only perform optimizations on $H(z^i)$, just in the same way as the control group established in section \ref{A:v}.
\paragraph{control groups 2} We design a restricted attention-network architecture by cutting off the links via $a$ between units, as shown in Figure \ref{fig:h2}.

Table \ref{table:sample-table} gives the scores of $z_t$, $c_t$ and $a_t$ before and after training.
The scores of the experimental group surpass those of its competitors.  

\begin{table}[ht]
  \caption{Scores of representations}
  \label{table:sample-table}
  \centering
  \begin{tabular}{llll}
    \toprule
             & $z_t$     & $c_t$ & $a _t$ \\
    \midrule
    Before Training & $2.04\pm 0.06$& $2.17\pm0.04$  &$0.29\pm 0.23$\\
    Experimental Group& \textcolor{red}{$3.13\pm 0.04$}& \textcolor{red}{$2.33\pm0.07$}&\textcolor{red}{$0.80\pm0.23$}      \\
    Control Group 1     & $2.93\pm 0.07 $&$1.81\pm0.23  $ & \\
    Control Group 2     & $2.93\pm 0.07$&$1.92\pm0.21  $& \\
    \bottomrule
  \end{tabular}
\end{table}    

\section{Recursive Modeling}
\label{RM}
Model building, arguably, is the approach to general intelligence\cite{lake2017building}.
Additionally, we think recursion is essential in the design of strong artificial intelligence, just as it is for many Turing complete machines\cite{turing1936computable}.
So we propose the approach of Recursive Modeling, which means that the agent should not only build causal models for the environment, but also recursively build causal models on the early-built ones.
The environment is where negentropy\cite{schrodinger1944life} flows in.

\begin{figure}[ht]
  \centering
  \includegraphics[width=0.7\textwidth]{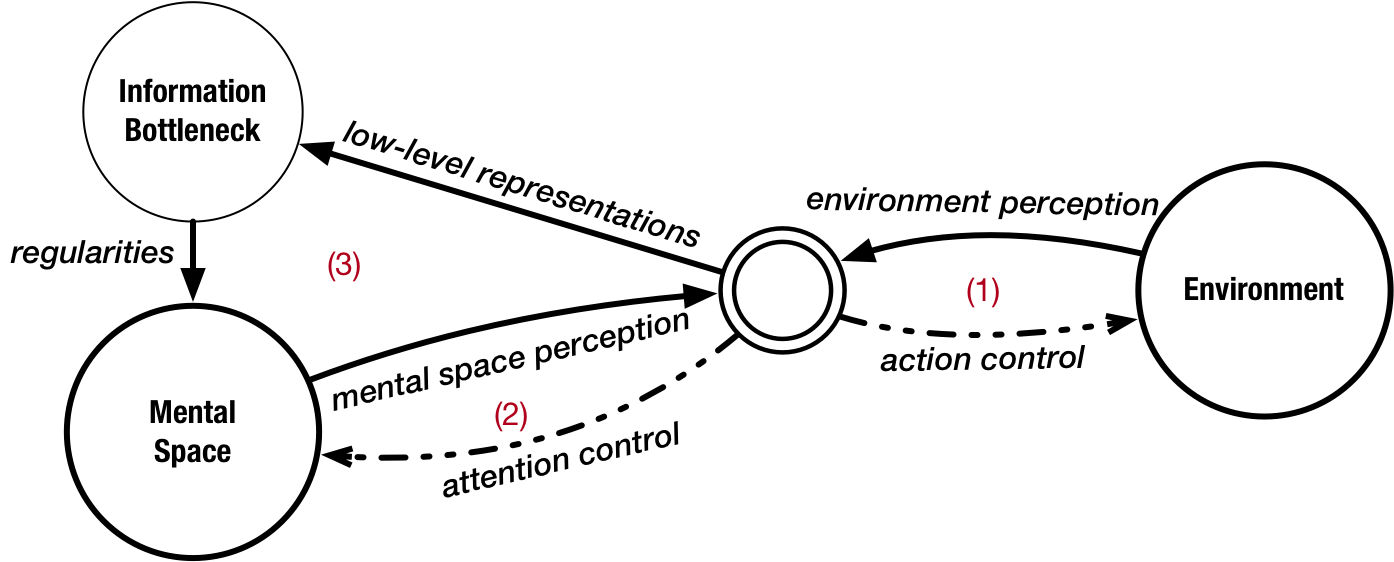}
  \caption{\label{fig:recursive_model} Schematic diagram of Recursive Modeling}
\end{figure}

As shown in the schematic diagram(Figure \ref{fig:recursive_model}), the Recursive Modeling approach has two requirements.
The first requirement is to build a mental space where models run.
If we think of a model as a collection of regularities (or schemas\cite{piaget1929child,bartlett1932remembering}), then the mental space is the collection of all regularities.
Regularities are usually obtained from information bottlenecks, like the linguistic regularities found in the word vector space\cite{mnih2013learning}, and the disentangled representations generated by generative models\cite{bengio2013representation,larsen2015autoencoding}.
Existing low-level representations should be recursively distilled by the information bottleneck.

The second requirement of Recursive Modeling is to allow the agent to perceive and intervene the mental space, just as it does with the environment in the physical world. Perception and intervention are two necessities to build causal models at any time.

Among the models that have been built, the early built models are to simulate the relations between real entities in the environment, while the later ones are responsible for abstract thinking tasks, such as calculus in a symbolic system.
We do not mean that there is a clear hierarchy between models.
In fact, the notion of "model" is only a fictitious concept describing a set of closely related regularities, and many of those regularities are actually intertwined and shared, and reappear at different levels.
Units in the cortex-network can also cluster into function regions, and regions can be organized in a hierarchy-like pattern.
Different models can correspond to different regions in the cortex.
However, this may be a future work and the article dose not involve this too much.

\subsection{BHN and Recursive Modeling}
\label{RM:bhn}
\begin{figure}[hbt]
  \centering
  \includegraphics[width=0.55\textwidth]{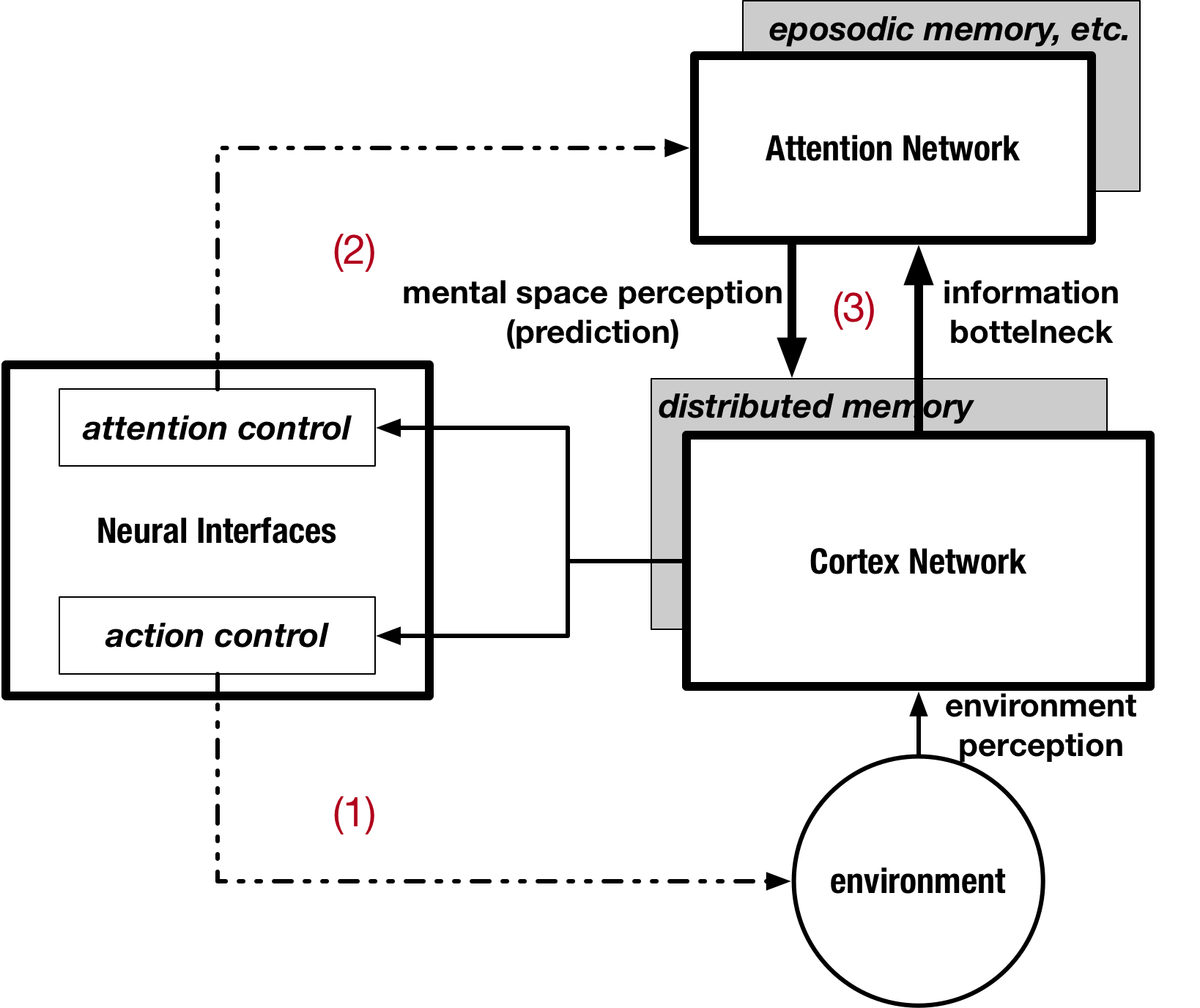}
  \caption{\label{fig:BHN} BHN adapted for Recursive Modeling}
\end{figure}

Figure \ref{fig:BHN} gives a schematic diagram of BHN adapted for Recursive Modeling, in which the three Loops marked in Figure \ref{fig:recursive_model} are also marked roughly at the corresponding positions.

BHN meets the two requirements of Recursive Modeling.
Firstly, the attention-network can serve as an information/attention bottleneck\cite{felleman1991distributed}, and the global attention($a$) can be regarded as representations in the mental space.
Secondly, it is possible for the agent to perceive the mental space by fusing bottom-up perceptions with top-down predictions, which will be detailed in the section \ref{RM:mpp}.

We think that much of the intelligence of the human brain resides in its sophisticated architecture, and now our BHN model is oversimplified and lacks many essential functions, such as dopaminergic neurons for reward and prediction error learning\cite{hollerman1998dopamine}, a realization of the attention control interface, the hippocampus forming mental maps and episodic memories, etc.
There is no doubt that we need more inspiration from the human brain to proceed with this work\cite{lake2017building}.

\subsection{Working Memory}
\label{RM:mpp}
We think that the human brain works by continuously mixing real perceptions with imaginary predictions, and in extreme cases it is like "hearing one's thoughts spoken out aloud"\cite{schneider1939psychischer}.
If $z_{t}^{i}$ represents what is heard, then the expectation $e^{i}_t
=
\sum_{z^{i}_j \in Z^{i}}w^{i}_{tj}
z^{i}_j$ can represent what the brain predicts to hear.
By replacing $z_{t}^{i}$ with $e_{t}^{i}$ at some times, the agent can somewhat perceive the mental space just as it perceives the external environment.

$z_{t}$ and $e_{t}$ are homologous, and they can both be used as the output of a unit, so that the information flow within the net is actually a mixture of perceptions($z_{t}$) and predictions($e_{t}$).
$z_{t}$ is involuntary and volatile, but $e_{t}$ is processed recurrently and remains somewhat locked inside the Loop {(\textcolor[RGB]{162,31,37}{3})}(marked both in Figure \ref{fig:recursive_model} and Figure \ref{fig:BHN}), and so in this way, $e_{t}$ can provide gain for $z_{t}$ in a sense.
We speculate that this mechanism corresponds to the brain's working memory, and its gain level determines whether the representations in the cortex will be suppressed or enhanced\cite{miller1991neural}.

\section{Experiment(2)}
\label{B}

\begin{figure}[b]
  \centering
  \includegraphics[width=0.7\textwidth]{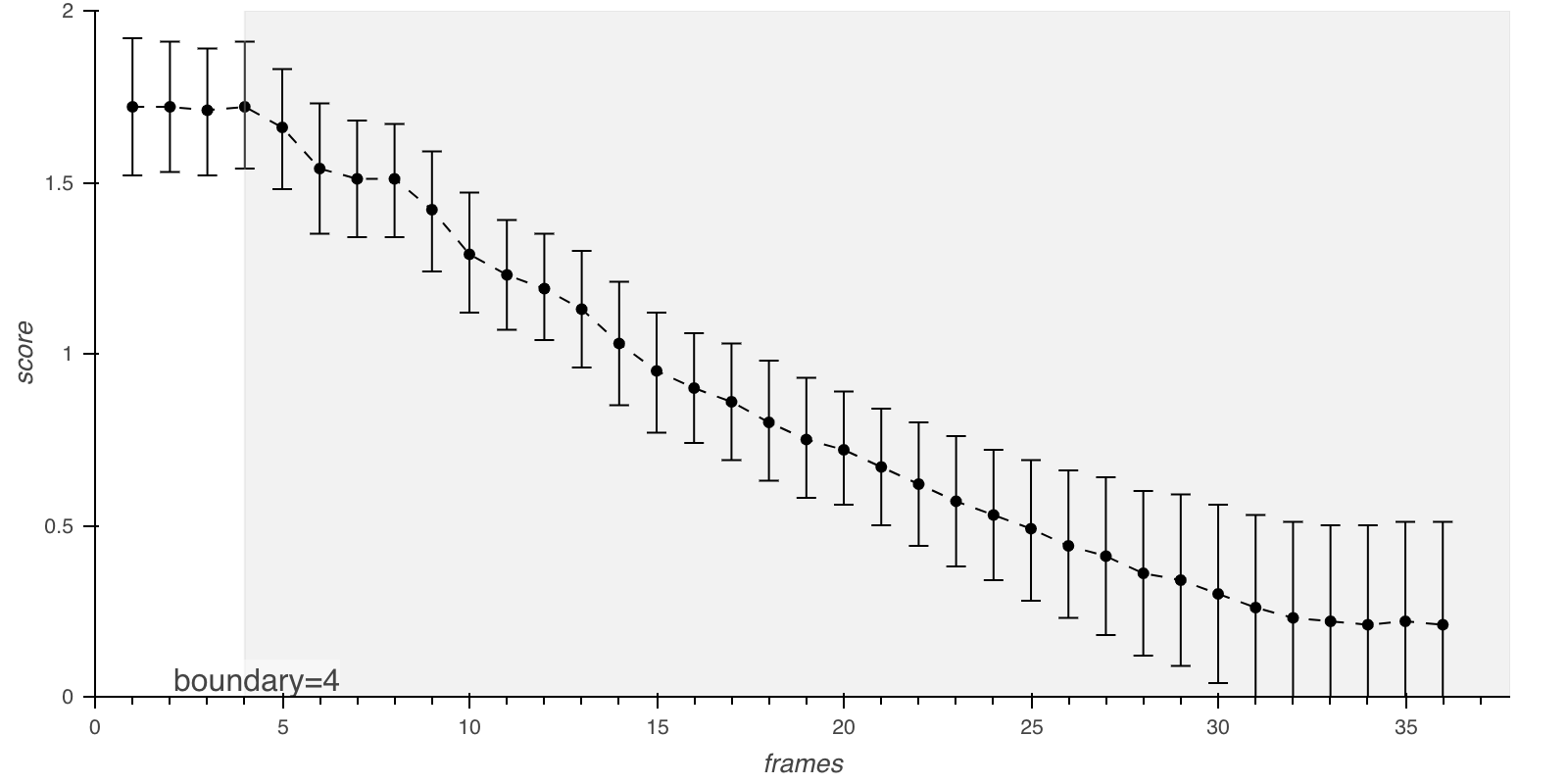}
  \caption{\label{fig:casual} The score of $e$ over time in the testing phase}
\end{figure}

We follow the same basic setup of the simple model in section \ref{A:v} to test the hypothesis of working memory by mixing $z_{t}^{i}$ with $e_{t-4}^{i}$.

First, in the training phase, we feed $z_t$ to $c^{i}_t = g^{i}_{ar}(*)$ for even time steps and feed $e_{t-4}$ for odd time steps.
A deconvolutional network reconstructing images from $e$ to give a score is also trained in this phase.

Next, in the testing phase, taking a certain time step as the boundary, $z_t$ is used before and $e_{t-4}$ is used after.
We judge the performance by how long the score of $e$ keeps positive in the testing phase.

Figure \ref{fig:casual} gives the result and it shows that the working memory effectively lasts for about 30 frames, much longer than the one frame which is what we adapt the system to in the earlier training phase.

\section{Conclusions}

In this article, we propose three hypothesises on the learning and working mechanism of the human brain. By formalizing these hypothesises, we get a computable objective, which is a sum of many objective functions. After that, we build and test a model(BHN), which couples several artificial neural networks together, to optimize the objective functions obtained. Finally, we propose the approach of Recursive Modeling and test a hypothesis on working memory.

\section*{Broader Impact}
Our work has no direct ethical or societal implications.

\bibliographystyle{apalike}
\small
\bibliography{reference}
\end{document}